# Causing is Achieving[1]

## - A solution to the problem of causation -

Riichiro Mizoguchi [a, b]

[a] *Japan Advanced Institute for Science and Technology (JAIST), Japan*
E-mail: *mizo@jaist.ac.jp*
[b] *Laboratory for Applied Ontology (LOA), ISTC-CNR, Trento, Italy*

Abstract:
From the standpoint of applied ontology, the problem of understanding and modeling causation has been recently challenged on the premise that causation is real. As a consequence, the following three results were obtained: (1) causation can be understood via the notion of systemic function; (2) any cause can be decomposed using only four subfunctions, namely *Achieves, Prevents, Allows* and *Disallows*; and (3) the last three subfunctions can be defined in terms of *Achieves* alone. It follows that the essence of causation lies in a single function, namely *Achieves*.

It remains to elucidate the nature of the *Achieves* function, which has been elaborated only partially in the previous work. In this paper, we first discuss a couple of underlying policies in the above-mentioned causal theory since these are useful in the discussion, then summarize the results obtained in the former paper, and finally reveal the nature of *Achieves* giving a complete solution to the problem of what causation is.

Keywords: Causation, systemic function, device ontology, direct causation, state-mediation, negative causation

## 1. Introduction

Although what causation is has been discussed for many years, it has been considered an unsolved problem to date. Assuming causation is real, this paper discusses causation as an occurrent (causing) and aims to demonstrate that it is essentially achieving. This view has been presented in [1] where the authors obtained the following results: (1) causation can be understood as a case of systemic function, and hence it is possible to talk about causation in terms of function, (2) Causing can be decomposed along two dimensions, direct/indirect and positive/negative, into the four subfunctions *Achieves, Prevents, Allows* and *Disallows,* and (3) *Prevents, Allows* and *Disallows* can be defined in terms of the *Achieves* function. It follows that causation is essentially understood in terms of *Achieves.*

The main goal of this paper is to provide a solution to what causation is rather than discussions about the existing theories of causation by uncovering what *Achieves* is. This issue has been only partially tackled in [1]. Before addressing the main topic, the paper reviews how an occurrent causes another. This review leads to fruitful discussions of causation which relies on three policies adopted in the theory: (1) the state-centric approach, (2) the distinction between direct and indirect causations, and (3) the impossibility of direct causation between events. On the basis of these observations and policies, we see why the conventional attempts to explain causation have not been successful. Then, after summarizing the results obtained in [1], we will turn to our main goal, the elucidation of the *Achieves* function. More precisely, we will reveal the nature of *Achieves* for all possible relata giving in this way a solution to the question of what causation is. Finally, some of the typical philosophical issues such as simultaneity, necessity and causal efficacy are discussed followed by conclusions.

## 2. Negative causation and Fundamental claims
### 2.1 Typical existing theories
The main questions driving this research can be articulated as follows:

(i) What is causation?
(ii) Does a theory of causation explain (exclude) the known positive (negative) examples?

---
[1] The title was suggested by Ludger Jansen.

Typical attempts to define the causation C→E are listed below [2]:

(1) E-type events regularly follow C-type events.
(2) E is counterfactually dependent on C.
(3) There is a spatiotemporally continuous transfer of energy or momentum from C to E.
(4) C raises the probability of E.
(5) By manipulating C, we could bring about changes in E.
(6) Dispositional account of causation [3][4].

Unfortunately, it has been shown that none of these successfully answers the above questions[2][5]. This is the reason why the survey of related literature is short. This paper is written intended to propose an innovative solution to the problem of what causation is rather than discussing about pros and cons of existing theories. Therefore, readers of this paper are expected to try to follow the innovative approach taken in the research and to capture the essentials of the solution that answers the above two questions.

**2.2 Negative causation**

Generally speaking, negative causation is considered particularly challenging. It is known that there are two types of negative causation: causation of absence and causation by absence. An example of the former is:

"No robbery occurred because John locked the door",

and an example of the latter is:

"A thief broke into Tom's house because he did not lock the door."

Causation related to absent occurrences is common in daily life, and hence causal theories that cannot explain them fail at least at the practical level. This said, it is difficult to talk about non-existing entities even in applied ontological systems, in particular those developed within a realist approach. While resolving this problem is a challenging task, the theory discussed in [1] can explain negative causation and overcome other disadvantages introduced in traditional approaches like, for instance, counterfactual theory [6].

**2.3 Fundamental claims**

Let us consider two examples: (a) Traffic accident and (b) Blood clot growth (See Fig. 3).

| | |
|---|---|
| a1. A car hit a person. | b1. A blood clot grew. |
| a2. A witness informed a fires station (EMS) of the accident. | b2. The blood vessel narrowed. |
| a3. An ambulance came and transferred the victim to a hospital. | b3. The flow rate of the blood reduced. |
| a4. The victim arrived at the hospital. | b4. An organ downstream was lacking oxygen. |
| a5. A doctor examined the victim. | b5. It contracted an ischemic disease |
| a6. etc. | b6. etc. |

In spite of the difference of the domain, these two are often considered similar in terms of causation because, so the argument goes, the causal chains they manifest are similar. However, a closer investigation reveals that they are very different from each other. What exists in the traffic accident is a sequence of completed processes (events). Each event happens after the former event has completed. On the other hand, at least the first three occurrents in the blood clot case happen concurrently, that is, it is wrong to say that the effective cross section of the blood vessel starts to decrease after the completion of the growth of the blood clot. This is a continuous and concurrent process with the decreasing in flow rate happening together with the clot growing. The occurrents in the case of the traffic accident have no direct interaction between them but interact with each other indirectly with the mediation of states, while there is no state-mediation in the case of blood clot and all the processes are ongoing even after the organism downstream has contracted an ischemic disease.

The issue here is whether the causal interaction between two processes is done by mediation of state or not. Therefore, the fact that these processes are constituted of states is irrelevant here. In fact, direct interaction between two processes is mainly done by equation-based parameter propagation where no state-mediation exists between them (See 5.2).

Note here that such ischemic disease will not be cured even if the growth of the clot has terminated, since it has been caused by the state of the smaller cross section of the blood vessel rather than the growth process itself. This fact tells us a very important message which is summarized as:

**Claim 1:** States play crucial roles in causation. So, states necessarily appear in the relata of causation.

Although all the interactions between the occurrents in the traffic accident happen with the mediation of states, those between b1, b2 and b3 in the blood clot case happen directly on each other, and hence they do not need any state in between the causal interaction to occur. This fact suggests a new distinction among causation kinds:

**Claim 2:** There are two types of causation: *direct causation* and *indirect causation*. The latter happens with the mediation of states, while the former needs no state-mediation in between.

Imagine two occurrents causally connected to each other. In the terminology of [7], if they overlap, then the interaction between them must be of the type holding between processes rather than between events[2]. Remember that any event, a completed process in [7], must be dealt with as a whole. This inherent property of an event prohibits direct interaction between events. This is obvious from the fact that no completed occurrents can directly influence another one after the completion. These observations lead to our third claim:

**Claim 3:** There is no direct causation between events (understood as completed processes).

These three claims challenge the traditional view of causation as found in the literature. The state-based approach to causation is the basis for the theory of causation discussed in this paper as we are going to see in the next section.

## 3. Fundamental issues for fruitful discussion on causation

In this section, we discuss two fundamental issues derived from the above three claims for better discussion on causation.

### 3.1 How an occurrent causes another

Suppose that two causally connected events/processes C and E occurred consecutively (in turn) in which C has influenced E in some way. Such an influence must be a kind of interaction between C and E. As to the question of when/where it happens, only the following four sequences are logically possible in which time passes from left to right and the symbol ";" denotes the separator between occurrents:

(a) C; the interaction; E
(b) C(including the interaction); E
(c) C; (including the interaction) E
(d) C(including the interaction); (including the interaction)E

The first case seems not possible because the interaction between C and E cannot happen at the time in which C had completed and E has not happened yet. Therefore, an answer can be provided only by the other three cases. The essence of these three is that the interaction between C and E are included in either C or E. Considering that any interaction requires the participation of the interacting entities, here C and E, it turns out that only (d) where C and E are overlapped is valid. This conclusion coincides with our claims discussed earlier. That is, the fact that the case (a) does not hold corresponds to Claim 3: "There is no direct causation between events". Case (d) corresponds the Claim 5 which will be introduced in section 5 and corresponds to: "Causal efficacy inheres in an event". Up to here, by interaction, we mean direct causation between processes or events. However, the case (a) turns out to be possible if we expand it to accept indirect interactions via states suggested by Claim 2. Some people might think an interaction, call it F, between C and E can be an ordinary process/event that mediates the influence of C on E. However, such thought fails under a regression argument because the same problem arises between C and F. Now let us consider the case where C and E meet (share the end and beginning instants). Direct causation is a positive causation so that a substantial interaction like energy or momentum transfer should happen for C to influence E. However, such transfer cannot happen in an instant time because there is no instant occurrent in reality. Therefore, (d) is valid only for overlapping cases.

In summary, (a) is possible only in the case of indirect causation mediated by states created by C, while only (d) is possible if we confine the discussion to direct (adjacent) causation. In other words, the secret of how an occurrent causes another is hidden in the case (d) which we are going to discuss in section 5.2. As far as the author knows, none of the existing theories on causation including the six theories presented in 2.1 except the first one cares about the directedness of causation. Although Humean theory discusses direct causation, it claims there is no causation in reality, which contradicts our theory proposed in this paper. Readers could consider that our theory of causation challenges Humean theory.

---

[2] See section 4 about definitions of events, processes and states.

### 3.2 Granularity and Adjacency

In the discussion on causation in the literature, not enough attention has been paid to the granularity of the occurrents and of the causal chain. In "John contracted cancer (E) because of smoking (C)", for example, a long causal chain is hidden between C and E, and hence these two events are not adjacent to each other in the causal chain. Special attention must be paid on the adjacency between the cause and its effect when discussing causation. Imagine, for example, a case where a discussion on the ontological nature of C → E is made in a causal sequence which, when better detailed, turns out to be C→$C^1$→$C^2$→$C^3$→E. The discussion would be problematic because it is not clear which causation is actually discussed. Is it the causation C→$C^1$ or $C^1$→$C^2$, or else? Indeed, it is unclear what one means by the causation C → E where C and E are not adjacent in the first place. Some might express objections by saying that it is theoretically difficult to talk about adjacency of causal occurrents because finer-grained causal sequences could exist between any causal pairs of events. Although the worry makes sense in many cases, it is not always true in reality. For example, "John kicked a ball (C) and it traveled away (E)" is a case where the cause and its effect are adjacent independently of the granularity. A kicking which is an impact-giving occurrent on an object by swinging one's leg is a complex of multiple micro motions, so it has a long causal sequence in it, but a kicking is a macro occurrent and refers to the entire sequence and hence the adjacency between the kicking and traveling occurrents remains unchanged. Concretely put, assume that a kicking (C) is composed of three finer causal units (say, $C^0$→$C^1$→$C^2$→$C^3$). Then the whole causal sequence becomes $C^0$→$C^1$→$C^2$→$C^3$→E in which the last event before E (the ball's traveling) is $C^3$, i.e., the collision between the foot and the ball. There is no finer-grained causal chain between C and E because C includes all events just before E happens. As described above, the kicking is thus a macro occurrent C that starts at $C^0$ and ends at $C^3$. The reason why the adjacency is not influenced in this example is because what can be decomposed by introducing finer-granular causation is only the causal occurrent itself rather than the causal link (→). The boundary between the cause and its effect under consideration remains unchanged in the case of decomposition of occurrents.

Another popular example: "John struck a match (C) and the match fired (E)", is also inappropriate for fruitful discussion of the nature of causation. There are several important causal occurrents between C and E, for example: "John struck a match (C) → frictional heat was generated → the gunpowder ignited → the match bar temperature rose → the match bar fired (E)". Because of the existence of such a finer causal chain in between, it is not appropriate to argue about the causal relation between C and E without a theory of composition of a causal chain.

The important issues about adjacency can be summarized as follows:

(1) Discussion of the nature of *causing* (denoted by →) using examples (C→E) must be done in the situation where the causal link being discussed is uniquely identified.

(2) In order to uncover the nature of causation, we should first tackle the adjacent causation before the nature of causal chains.

(3) Our new finding on directedness of causation is of importance because it is compliant with (1). Any direct causation has its unique "*causing* (→)" in C→E and problematic issues appear only in indirect causations.

(4) Direct causations are unaffected by additive/subtractive interventions because, in the case of <Event → State> pattern, when the happening of the effect (E) is under consideration, the causal event (C) has completed already which guarantees E has happened, and in the case of <Process → Process>, as is discussed in 5.2.2, both C and E are happening concurrently and hence at the time E exists C is necessarily happening. That is, there is no way to separate them. Needless to say, if C does not happen, there is no need to talk about E's occurrence. (In the case of indirect causation and/or a long causal chain, however, an intervention could occur at any time after C has completed)

None of the six existing theories except the first shown in 2.1 cares about the adjacency of causation.

### 4. Summary of the functional theory of causation

This section summarizes the achievements obtained in [1], which will be extended in section 5 to complete a presentation of the entire theory. First of all, we present informal definitions of state, process and event.

A **process** is an ongoing/progressive occurrent which exists at any instant as a whole while it is activated. It constitutes an event when completion.

An **event** is a completed occurrent and is dealt with as a whole spreading over the entire interval. It has temporal parts. It cannot change, while a process can. A sequence of events can form a process.

A **state** is a time-indexed quality and its consecutive change constitutes a process. While the quality inheres in the bearer, it (the bearer) participates in the state when viewed from temporal perspective.

### 4.1 Causation and function

Given a causation C→E, an entity A composed of all entities participating in C can be said to perform a systemic function (see [8]) that achieves E by behaving in a certain way in C. Indeed, considering E as a systemic goal for C, we obtain the following claim:

**Claim 4:** Causation can be mapped to systemic function.

Technically, the validity of Claim 4 relies on a series of results that led to the introduction of notions like systemic context and systemic function [8], and theories like the device ontology [9]. The device ontology is a system of roles that generalizes an entity that acts as a device in the propagation of other entities (e.g., the heart as a device in a human body), and enables us to talk about behaviors of an entity in a system and of externally detectable input/output characteristics. The modeling is granularity-independent thanks to the recursive application to behavioral analysis of the components where necessary. Although we will briefly explain it in 5.1, readers can find the details of the device ontology in [1] and [9].

### 4.2 Essentials of causation

Note first that Claim 4 does not have a revisionist goal, it does not attempt to free the world from causation. The mapping instead aims to shed light on causation by studying it in terms of functions. In other words, while causation is assumed to be a truly ontological notion (Presupposition 1), its rewriting in terms of function helps to deepening its understanding. Of course, there is an obvious circularity in the causation-function mapping due to the fact that causation inheres in function. However, this circularity does not raise a problem here since there is no attempt to reduce the former to the latter. On the contrary, since the circularity makes evident the strict dependence of function over causation and since we already have a well-developed theory of function, it is interesting to see what we can learn by analyzing causality in terms of functions. The core of a function is the achievement of a goal by a behavior, which suggests that the most generic function is the *Achieve* function as we are going to show.

### 4.3 The functional square

As stated earlier, there are two distinctions about causation: direct/indirect and positive/negative. *Achieves* is direct and positive. According to these two dimensions, there are four types of functions, see Table 1, with *Achieves* being classified as direct and positive.

*Achieves* is investigated in section 5, here we define the other three.

*Prevents* is defined as follows (for details see [1])

**Definition 1** *(Prevents):* In context C, an occurrent X *prevents* an occurrent Y if and only if there exists an occurrent Z in C such that (i) X *Achieves* Z in C, and (ii) Y and Z are incompatible in C by which we mean either Y or Z is true but not both in C.

Table 1  Functional square.

|  | positive | negative |
|---|---|---|
| direct | Achieves | Prevents |
| indirect | Allows | Disallows |

Some readers may find it uncomfortable to bring up context when discussing causation, but this is by no means a relativization of causation to context, rather the opposite. It is by selecting a causation that automatically one also determines a context in which the causation holds. Imagine, for example, that "John could not enter the room because the door was locked". In order to make this causation valid, a context is required to ensure that there is no other way to enter the room (e.g., there is no other unlocked door or window) and he is not allowed to break in. Causal talks in the literature implicitly assume the existence of such contexts.

### 4.4 Indirect causation and preconditions

Many of the examples discussed thus far in the literature are indirect causation as shown below.

Example 1: No robbery occurred because John locked the door.

Example 2: A thief broke into Tom's house because he did not lock the door.
Example 3: Emily's dog was bitten by an insect, contracting an eye disease as a result. Ignoring this, she did not take the dog to any animal hospital. Later the dog lost its sight. (from [10]).
Example 4: A father pushing his child out of the way of a speeding car saved the child's life (i.e. causes it not to die). [11].
Example 5: The failure of delivering a piece of machinery in time caused a machine to break down.[12]
Example 6 (Double prevention): Suzy's plane will bomb an enemy target if she is not shot down by an enemy pilot. Piloting another plane, Billy shoots down the enemy pilot. As a result, Suzy bombs the enemy target. "Suzy bombed the enemy target because Billy shot the enemy pilot." [16]

Nevertheless, the discussion of causation in the literature has been based on one type of causation "An occurrent C caused an occurrent E" without distinguishing between direct and indirect causation [2][5]. Our proposal is to adopt a state-centered view and to conceptualize indirect causation as state-mediated causation. From [1] we know that the result is significant. Taking this approach in combination with the functional view, one can show that many examples discussed in the literature are not about the *Achieves* function, but about the *Allows* and *Disallows* functions. In fact, none of the above examples is characterized by *Achieves, but by (1) Disallows, (2) Allows, (3) Allows, (4) Disallows, (5) Allows and (6) Allows (double prevention)*, respectively, as shown below:

(1) John locking the door Achieves the locked state of the door, which Disallows a robbery.
(2) The unlocked state has been Maintained (by Tom's not locking the door), which Allows the thief's break-in.
(3) The state where Emily's dog stays in her house has been Maintained (by her not taking the dog to a hospital), which Disallows her dog to be cured from its eye disease, which Allows it to lose its sight.
(4) A father pushing his son Achieves the state where he is out of the course of a speeding car, which Disallows his son to die by being hit by the car.
(5) An unknown event Disallows a state where a piece of a machinery is available by Preventing the success of the delivery in time, which Disallows the recovery of the machine by fixation, which Allows the machine to break down.
(6) Billy shooting down an enemy pilot Disallows the enemy (who would have shot down Suzy's plane to Disallows her to bomb an enemy target) to shoot down Suzy's plane, which Allows her to bomb an enemy target.

### 4.4.1 Preconditions
A precondition for an occurrent is a condition whose truth value influences the activation of the associated occurrent. Before defining the latter two functions, we note that there are two types of preconditions for an occurrent. We call one the *facilitative* precondition and the other the *preventive* precondition. Either one can be stated as follows:

**Facilitative precondition for X:**
If occurrent X occurs in context C, the facilitative precondition of X is true in C.
**Preventive precondition for X:**
If occurrent X occurs in context C, the preventive precondition of X is false in C.

### 4.4.2 *Allows* and *Disallows*
We can now define the functions *Allows* and *Disallows*.

**Definition 2 (*Allows*)**: Occurrent X **Allows** occurrent Y in context C if and only if there is a state Z that satisfies one of the following conditions: In C, (i) X *Achieves* Z and Z is a facilitative precondition for Y, (ii) X *Prevents* Z and Z is a preventive precondition for Y, and (iii) X maintains Z and Z is a facilitative precondition for Y.
**Definition 3 (*Disallows*)**: Occurrent X **Disallows** occurrent Y in context C if and only if there is a state Z that satisfies one of the following conditions: In C, (i) X *Achieves* Z and Z is a preventive precondition for Y, (ii) X *Prevents* Z and Z is a facilitative precondition for Y, and (iii) X maintains Z and Z is a preventive precondition for Y.

Note that the last two definitions use the function *Maintain* which originally refers to the function of outputting the incoming operand without changing it. The prototypical example is a conduit that outputs exactly what is fed into it: a water pipe has an output flow rate that is equal to the input flow rate. The general *Maintain* function performs similarly by keeping the current state unchanged. In terms of *Maintain,*

"John did not lock the door" is paraphrased as he *Maintained* the lock in the unlocked state. In other words, the omission of an event that should have occurred *Maintains* the current state (by doing nothing). The introduction of *Maintain* thus overcomes the theoretical difficulty of mentioning an occurrent which does not happen.

## 5. On the *Achieves* function

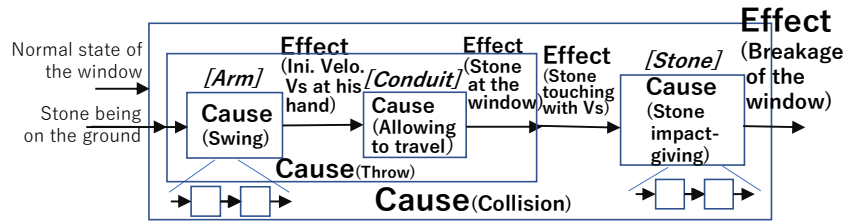

Fig. 1 Configuration of the window-breaking system, in which all inputs and outputs are states and all that are written headed by "Cause" in the box are events. In the finer granularity, some causation are of type Process => Process such as arm-swinging => stone-flying. The releasing event is omitted.

The discussion we have done thus far enables us to confine the study of the meaning of causation to that of *Achieves*, and hence we need to investigate this function to shed light on what *causing* is. This section is devoted for this purpose.

### 5.1 Rule of how to apply the device ontology

The device ontology [9] is a system of roles such as *device (actor), conduit, operand* and *medium* for capturing dynamic systems in a consistent manner. A device receives *operand* carried by *medium* and processes it to output. It has a nested structure for finer-granular phenomena. It is used to define function in a systemic context in which its input-output behavior realized by the internal processing plays the functional role under consideration in the given systemic context [8]. Assume a causation C → E is given. Considering E is a systemic goal achieved by the occurrent C, an entity composed of all objects participating in C can be said that it performs a systemic function that achieves E using C in that context. This conceptual framework is elaborated in terms of "what to achieve" and "how to achieve" defined in function decomposition proposed in [13]. Then, we have the following: "what to achieve" corresponds to E and "how to achieve" to C, which shows a causal sequence of C→E is equivalent to *Achieves* function performed by the device identified by the above way. The implication of this device ontology-based modeling is significant. The key idea is to consider C in C→E an internal phenomena (occurrent) happening in a device which performs *Achieves* function. It contrasts the ordinary notion of a causal chain which is conceptualized as a sequence of causally connected occurrents at the same level. On the other hand, our modeling is nested in multiple devices of different granularity in such a way that a cause is one-level deeper than the corresponding effect as is depicted in Fig. 1. Thanks to the granularity-free characteristic, the above way of application of device ontology to modeling of causation works for any causal pair of occurrents.

For example, suppose a situation where Tom threw a stone to the window of a house and it was broken because the stone hit it. Before the stone-throwing event, it seems there is no system which has a systemic function there. Tom, the stone and the window seem to exist independently of each other without forming any system and such a system does not have to be there. However, a system emerges when he has thrown a stone to the window. That is, whenever and wherever C→E exists a system is there because of the identification of a causation which reveals a certain connection between participants of the related occurrents. At the time when Tom held a stone and started to throw it to the window, his arm/hand, the stone and the window virtually form a system (device) in which three behaviors (occurrents) happen in turn: throwing motion, stone traveling and the stone hitting the window. Whether or not Tom has an intention to break the window does not matter here. Once a causation "Tom threw a stone to the window of a house and it was broken" is picked out, we can talk about a window-breaking system composed of Tom's arm/hand and the stone together with a window as a target object. There would be three-level granularity. The coarsest granularity suggests to model the phenomena as a system (device) composed of the stone and his arm whose input is the normal state of the window and output is the breakage of the window caused by the collision, in which all the three occurrents are included (See Fig. 1). What happened in this system is analyzed in the second coarsest granularity. That is, two subsystems are identified: (1) the stone as a window-hitting subsystem and (2) his arm and a virtual conduit (the 3D region occupied by the trajectory of the stone) with the stone as a stone-travelling subsystem. In the first subsystem, the stone touching (hitting) the window with velocity Vs corresponds to the cause and the breakage of the window to the resultant occurrent (effect). The second subsystem models how the stone touches the window with Vs with input of the initial location (ground) of the stone and output of the final location (at the window) with Vs. The resulting configuration of system/subsystems is shown in Fig. 1. We now obtain a rule for the application of the device ontology in causation as occurrent:

**[Rule]** Identify a device by assigning E as output (what to achieve) and C as the internal behavior (how to achieve). If necessary, apply this rule to C recursively (e.g., in the case where C is a causal chain made of finer-granular occurrents).

### 5.2 The essentials of *Achieves*

Finally, it is time to investigate the *Achieves* function in all cases of relata in terms of the possible types such as event, process and state.

#### 5.2.1 Event → Process

A typical example is a collision of two balls in a billiard game. Imagine that a rolling white ball collides with a red ball at rest with velocity *Va*, and the red ball rolls afterwards. At first glance, it seems that the collision event causes the red ball rolling, but this is not accurate. There is no instant event in reality and any collision is a quick push process with a large acceleration. Because it is incorrect to say that the red ball starts rolling after collision, this should be <Process → Process> causation rather than <Event → Process>. In addition, what is activated by the collision is not an event but a process. Although some people might think the red ball rolling must begin by a starting event, but such a starting event is essentially a sub-event associated with all occurrents, it cannot be a relata of causation, since any relata must be an entity as a whole rather than a part. Another seemingly problematic example which seems to belong to the <Event → Process> pattern would be <A switch was turned on → the machine worked>. However, because what caused the machine to work is not the switch-turning behavior but the resulting "on" state of the switch, this is not a direct but an indirect causation. The above observation suggests that there is no <Event → Process> pattern in the real world. This is due to the nature of an event which must be dealt with as a whole at any time [7] and once it has terminated there is no way to influence others except via resultant intermediate states. However, such state-mediated causation is an indirect causation. If the causal influence happens before the termination, then it is not an event but a process which performs the influence as discussed in the next sub-section.

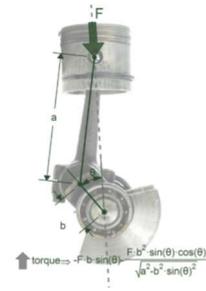

Fig. 2  Pair of a piston and a crank.

#### 5.2.2 Process → Process

A typical example of this pattern is a pair of a piston and a crank (see Fig. 2). The piston and the crank are connected to each other by a joint so that their behaviors (piston's downward and crank's rotation) are interlocked and are necessarily synchronized. In other words, the two processes happen concurrently which is the common interaction of this pattern. A remarkable feature of the P→P pattern is that its effect is generated via laws such as *f=ma* or law-like formula such as *x=y* which is free from causation. In the case of the push process of the piston & crank, for example, it seems that the piston exerts force *f* on the crank and the crank moves with acceleration *a* (= *f/m*). Although it would be correct in terms of Newtonian mechanics, we try to avoid application of physical laws as much as possible because they are hypotheses in nature and they might unconsciously introduce hidden causation. Therefore, we concentrate on the fact that there exists a shared individual in any pushing/pulling phenomena caused by contact force. That is, we identify an equation *x=y* in which x and y are positions/speeds/accelerations of the bottom edge of the piston rod and the edge of the crank, respectively. We talk of these two behaviors (motions) without using formula *f=ma*. (The equation *x=y* is not a law but a causation-free formula which holds a priori and hence we can accept it as given truth.) This is why we introduce *equation-based parameter propagation* discussed below. A similar equation is found in the blood clot case shown in Fig. 3 in which *x=y+z* holds where *x* is the cross section of the blood vessel, *y* is the area of the vertical cross section of the clot and *z* the cross section of the effective channel of the blood vessel. This is also such a formula that we can accept as given truth rather than an equation derived from a physical law. A typical example of *Achieves* of this type is *throwing (an object)* as a process in which the position and velocity is shared by the hand and the object ($x=y$) and *Allows,* by releasing, the object to travel with the initial position and velocity at the time of the release. Other examples include *push*, *pull*, *warm*, *cool, etc.*

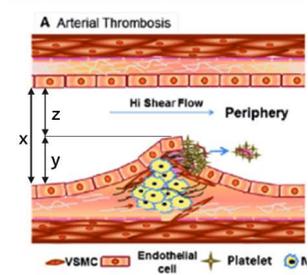

Fig. 3  Blood clot.

**(a)** *Equation-based parameter propagation*
The issue we need to discuss as to the causation in the pattern of P→P is whether the propagation through such formula is causal or not. Although y increases if x increases in the case of $x=y$, it happens not because of the relation between x and y is causal, since the reason why $x=y$ holds in the case of the piston & crank is because both objects share the same entity (the joint) and there is only one motion of the same entity with two different perspectives (the piston and the crank perspectives). So, it seems there is no causation in the propagation made through $x=y$. However, there exist two phenomena in which the piston pushes down the crank and the crank pulls down the piston. These two phenomena are causally different, though the observable behaviors are the same. We need a more careful investigation into this.

We have neglected <Process → State> as a pattern in the discussion because it is understood to be included in the pattern of <Process → Process> because a process is a continuous/consecutive change of states. Nevertheless, if we try to understand what is happening at each instant during the interval when the process is ongoing, it is useful to investigate the <Process → State> case. Similar to the discussion in the case of <Event → State> discussed in 5.2.4 in which the state is obtained as the result of the completed event, the state on the right-hand side of <Process → State> seems to be also the result of the process on the left-hand side. To support this observation, we need to understand the state in the process and the propagation from the state on the left-hand side to that on the right-hand side. The conclusion is the state change of the left-hand side is not caused by the process because it is another description of the same process which is the change of the state but that of the right-hand side is causal because the state is created by the process as the case of <Event → State>.

Note here that a process is a sequence of state change. In other words, a state or a set of states form a process when they change as time goes, and as a result, a new state keeps being created at any time while the process is ongoing. By the above conclusion, we do not mean any process cannot causally generates a new state. All processes can be said that they have causation in their state change. Let us examine this claim for the possible two cases:

[Case 1] In the case that the process under consideration has an operand (target object), the state change of the operand is the result of some operation done by the action/occurrent intrinsically specified in it. So, the state change of the object is causal.

[Case 2] In the case of no operand, that is, occurrents denoted by intransitive verbs, on the other hand, we need more neat consideration.

Let us take "growing" as an example. Growth of an organism is made thanks to a lot of internal biological activities of its parts. The result of the growth is of variety according to attributes/states we pay attention. When we concentrate on its size, we observe enlarging process of its size and it is a sequence of state change in terms of the size. Apparently, it is caused by those internal biological activities in the growth. Although walking is an action denoted by an intransitive verb *walk*, <John walking → John's location changing> is causal because walking has a couple of complex internal causal occurrents inside which enables John to walk and his location change is its effect.

The above discussion seems to have contradictory conclusions: *A process does not causally generate a state vs. a process causally generates a new state*. This apparent inconsistency derives from what we mean by "process". In the example of walking process, it is true that John's *walking* process causally generates a new state: where John is. However, if we mean the John's *moving* process, the conclusion is different because *moving* is a description of John's location change with no causal implications, that is, how the location change is realized is missing.

### 5.2.3 State → State

We have discussed <Event → Process>, <Process → Process> and <Process → State> patterns. The next is the pattern <State → State>. The blood clot case shown in Fig. 3 is a typical example of this pattern which is analogous to a flow-control valve in fluid dynamics in which <The half-closed state of a valve → The half flow rate> holds. In general, such a case where a state can be a cause of some occurrent is limited to the case where it is an internal state of a device. Note that the growth of a blood clot is equivalent to dynamic generation of a flow-control valve (with closing behavior) in the blood vessel.

Causation we have discussed thus far is based on change characterized as the non-zero derivative and on its propagation. In particular, device ontology works mainly for such changes. Note here that there is another kind of change in the talk of causation. It is the deviation from the standard and typical examples are found in flow-control valve and the blood clot cases discussed above. We do not have a causal theory for such change yet. The causation < the half-closed state of a valve (S1)[3] → the half flow rate (S2) > is not based on the non-zero derivative changes and is a complex causal phenomenon based on finer-grained causal occurrents. Although it seems difficult to tackle it at first glance, it is fairly easy to model it in terms of the <Process → State> type causation together with the recursive application of device ontology based on the rule discussed in 5.1.

Let us take a flow-control valve as an example. Assume the valve has started to close and stopped at an intermediate state between full open and full closed. Let S1 be the valve opening state after the deviation from the full open state. The phenomenon as a whole is a closing event E1 but <E1 → S1> is not valid because it is not causation but a description of a phenomenon from two different perspectives. Nevertheless, there are several internal phenomena which causes to increase the resistance against the smooth flow of the fluid and hence the flow rate is smaller than the standard (S2). To see what happens in terms of changes with non-zero derivative, we investigate it according to two cases: (i) valve's closing process P1 is ongoing and (ii) P1 has terminated at the half-closing state. Following the theory in [7], the closing event (E1) is constituted of P1. While P1 is ongoing, the valve closing process P1 causes increase of the resistance against the smooth flow of the fluid so that the flow-rate-decreasing process P2 is ongoing. Therefore, <P1 → P2> holds and at any time instance during the process is ongoing <P1 → S2> holds. On the other hand, when P1 has terminated, then <E1 → S2> is established and S2 is output. Considering S1 is also the state when E1 has terminated, the above observations suggest that the pattern <S1 → S2> does not mean propagation of changes but rather continuation of causal phenomena caused by an occurrent of other patterns, say, <P1 → P2>. We do not have functions of this type other than *Cause*. In terms of Adjacency discussed in 3.2, S1 and S2 are adjacent to each other because the refinement of the granularity does not change the boundary between these two. Note here that <Process → Process> stands for transitional situations of causal phenomena, while <State → State> stable situations.

It would be worth to note the fact that both <E1 → S2> and <P1 → S2> where E1 is constituted by P1 hold in the case of flow-control valve is neither usual nor trivial. The continuity of the output state between an event and its constituent process does not always happen. Imagine a cutting process and the corresponding event. At any time when the cutting process is ongoing, the object has not been separated yet. It has been separated only after the completion of the cutting event.

### 5.2.4 Event → State

The last pattern is <Event → State>. Whenever an event has completed, a resulting state is generated and it lasts for a while. When you cut an object, it becomes separated. When you buy a good of 10 USD, then your pocket money decreases by that amount and you obtain the ownership of the good. When you finish your daily walk to the park, then you are back to your home again. In all cases, the corresponding resultant states are successfully realized or achieved when any event has completed. Note here that in such cases, no intervention would occur before getting the resulting state. This is because if the event has not completed, there is no need to talk about its effect. As we see thus far, this pattern nicely demonstrates that the essential of causation is generating a new resultant state which would eventually causes (*Allows or Disallows*) succeeding occurrents. Considering <Process → Process> and <State → State> patterns tend to work as continuation or maintaining of existing states and/or their change [14], <Event → State> pattern plays the key role in causation by creating a new state. As we have already seen thus far, there are many example functions of this type and *cut* is one of the most typical ones. Considering that most processes will have eventually terminated to constitute an event with resultant states, many of the concrete functions belong to this type such as *kill, produce, inform, delete, transfer, join, divide, etc.*

### 5.3 How to terminate the recursive application of device ontology

In the above discussion, especially in the case of <P → P>, analysis of the internal causal phenomena is deferred to recursive application of device ontology to cope with finer-granular phenomena as indicated by the rule presented in 5.1 (See Fig. 1). Completion of our discussion on causation requires us to guarantee the recursive application will terminate without leaving any ambiguity of what *Achieves* is. This problem is resolved as follows.

---

[3] Hereafter, S1 and S2 denote not only the state "half-xyz" but also any state deviated from the standard value (state).

Recursive application of the device ontology definitely terminates when it reaches the minimal granularity. So, we have the following two cases:

(a) The change under consideration cannot be attributed to the inner causal phenomena of the device, and hence it is because of proactive phenomena such as Alpha collapse which are not reduced to more fundamental phenomena.
(b) Except (a) there are only two cases: either <E1→S2> or <P1 → P2/S2>. In either case, S2 has been obtained using a certain propagation of S1 causally obtained by E1 or P1.

The latter is apparently what we have discussed thus far so that we can identify causal source. Only the case (a) remains to be discussed. The issue is how many such principles that are irreducible to finer causal phenomena, exist and how much they influence our theory of causation. Although we cannot provide a proof, there seems to exist only a few such principles, which we introduce as primitives into our theory. We can summarize what we have discussed in the following three claims:

**Claim 5:** Causal efficacy inheres in each event as the pattern of <Event → State> rather than between events. It is production of a new resultant state when an event has completed.
**Claim 6:** *Achieves,* which is the core subfunction of causation, is characterized by the three patterns: <Event → State>, <Process → Process> and <State → State>.
**Claim 7:** Any causation in reality is a sequence of occurrents connected by either one of the four subfunctions: *Achieves, Prevents, Allows* or *Disallows*.

We need to elaborate the role of *Allows* in causal chains. Informally put, it makes one of the facilitative preconditions of an occurrent true so that it literally allows the occurrent to happen, that is, the occurrent does not necessarily happen then and when it happens is undetermined.

## 6. Discussion
In this section, we discuss several philosophical issues.

**Direct and indirect causation**
We have extensively discussed the secrete of causation and claimed it lies in direct, that is adjacent causation rather than indirect causation. Unfortunately, however, indirect causation has been mainly discussed in the literature to date without paying attention to the directness (adjacency) of causation. This distinction is the source of the success of our theory as elaborated below.

**Simultaneity**
Most people believe that the cause must have happened before the effect. One of the remarkable findings of our theory is that it is only true in the case of indirect causations in which states exist between the cause and the effect to mediate the causal influence from the cause to the effect. In the case of direct causation, however, there exist simultaneous causations as we saw in the <Process → Process> pattern of causation discussed in 5.2.2. The causal influence from the piston to the crank occurs at the same time because the two processes occur concurrently due to the fact that these two objects share the joint as their part. In the case of mechanical causations, there exist quite a few examples of simultaneous causation. Although it is counter-intuitive, we have to accept this true characteristic of direct causation. The reason why it is hard for us to accept this fact would be mainly because we often encounter indirect causations in daily life and get used to the non-instantaneous influence from the cause to the effect.

**Necessity**
The typical direct causation, *Achieves* is necessary, while *Allows* is not. As we have discussed above, *Achieves* of the <Event → State> type is necessary because the state as the effect is obtained at the time of the completion of the causal event. Because most of the real causations are such causal chains that include *Allows* subfunctions, we can conclude they cannot be necessary. It is because *Allows* just makes one of the facilitative preconditions of the effect true. Although the argument is so simple, the long-lasting debate about the necessity of causation has been resolved by our theory.

**Sufficient conditions for an occurrent**

Another topic related to necessity is what a sufficient condition is for an occurrent in reality. Contrary to logic, it is understood that there is no sufficient condition for any occurrent to happen in reality. Imagine a simple example of pushing an object. According to Newton's second law: *f=ma*, when a force *f* is exerted on an object whose mass is *m*, it moves with acceleration *a*. Although it is theoretically true, there exist a lot of preconditions that have to be satisfied for the object actually to move in reality. Examples include *f* is larger than the friction force, the object is not blocked by, say, a wall, the pushing device can generate *f* without being broken, the object is hard enough to accept *f* properly, etc. Although people believe that exerting *f* is the sufficient condition for moving the object, it is not the case in reality. First of all, all the facilitative preconditions must be true and all the preventive preconditions must be false. If not, exerting *f* cannot cause the object to move. Even worse, assume a situation where *f* is being exerted, and all of the facilitative preconditions are true and all the preventive preconditions are false except the blocking by the wall. In such a situation, the object remains still. If the last true preventive precondition has become false by removing the wall, then the object starts to move with acceleration, which tells us as if making the preventive precondition false is a sufficient condition for the move of an object. Of course, it is not the case. The truth is as follows.

First of all, exerting a force should be understood as one of the facilitative preconditions for the move of an object. Then, what exists, in reality, is an Achievement of the last satisfaction of a facilitative precondition or the last falsification of the preventive precondition among all the preconditions can play the role of the sufficient condition of the occurrent under consideration. Even if exerting a force f provides the target object with motion energy, it does not always play the role of a sufficient condition for the object to move. Assume a simple case where no preventive preconditions, when an Allows happens to make the last false facilitative precondition true, the occurrent happens at the very time when *Allows* has happened. A typical example is a switch-turning event which is understood as an *Allows* for a machine to work by Achieve-ing the "on state" of the switch which must have been in the last false facilitative precondition of the machine to work.

**The main body of causation**
Claim 4 can be paraphrased as "The main body of causation is a systemic function". The conventional investigation of causation/causality is based on a presupposition that causation cannot be observed because there seems to exist no tangible occurrent corresponding to causation, which might have led to that causation/causality has been dealt with as a relation between occurrents rather than an occurrent. One of the significant contributions of our research on causation would be we have successfully captured the main body of causation as a systemic function. A concrete example of a causal occurrent is a cutting function which is an instance of *Achieves* of <Event → State> pattern. In other words, a cutting function realizes causation <cutting event → being separated> (see 5.2.4) that is perfectly observable.

**Causal efficacy**
There is a causal theory that claims causal efficacy lies in an object as a disposition [4][5]. It is unique in that it tries to explain causation in terms of the more primitive notion that is disposition. A typical example is the fragility of a glass that shatters when a stroke, called a trigger, is given. According to the theory, the shattering is considered as the manifestation of the disposition, fragility, of the glass. It claims disposition has the power to invoke a causal phenomenon, and hence the causal efficacy lies in disposition. It is attractive. However, it needs to carefully deal with a trigger that is necessary for the manifestation of the disposition because one might be able to say the trigger caused the manifestation. In other words, one has to defend the theory against such a criticism that suggests the necessity of the explanation of how the trigger invokes the manifestation of the disposition because it seems to be a kind of causation.

It is interesting to find a similarity between the dispositional theory of causation and our theory. Although both claim causal efficacy lies in an entity, the former claims it lies in an object as a disposition, while the latter in an occurrent. According to our theory, the glass shattering is explained as follows: the stroke as a trigger generates an impact. If the impact is small enough, the glass does not break but moves, while if it is larger than a threshold, the glass shatters into pieces that *Allows* someone to hurt by generating a dangerous state of the floor. How the impact causes *(Achieves)* the glass to break (shatter) is explained by <Process→ Process> pattern because it is a very rapid and strong push. At the finer granular level, we can see the first crack is generated and it propagates in many directions, etc.

7. **Concluding remarks**

We have discussed causation not as a relation but as an occurrent at the token level and claimed that ***causing corresponds to achieving***. After presenting three main claims we discussed how an occurrent causes another together with the role of the notion of directedness of causation. Then, we gave a brief summary of the achievements obtained in [1]. We finally investigated the nature of the direct function *Achieves*, and showed that the source of causation inheres in an event and that the <Event → State>, <Process→ Process> and <State → State> patterns play the key role in causation.

We learned the investigation into indirect causations such as *"John struck a match and the match fired"* does not lead to a successful investigation into the nature of causation because such indirect causations are composed of multiple direct causations each of which bears the nature of causation in it.

Although we have introduced the functional perspectives to causal talk, we are not trapped so-called a circular discussion due to the causal nature of a function because we do not use the causal nature of a function to investigate causation. Instead, we have investigated the function Achieve in terms of types of arguments and found these two patterns <Event → State> and <Process→ Process> are essentials of causation in the sense that they bear primitive causation in themselves[4].

We have outlined a proposed theory of causation. The future work includes proper positioning of the theory in the rich history of causal theories. Apparently, such a causal chain that is composed only of *Achieves* would be equivalent with *Achieves* as a whole. However, many of the real causal chains would be composed of the four subfunctions: *Achieves, Prevents, Allows* and *Disallows*. We need a theory for explaining what a causal chain is as a whole in terms of these for subfunctions. A rigorous definition of system and state, on which the theory relies, is still lacking. These are genuine metaphysical issues and hence go beyond the scope of this paper.

**Acknowledgement**
The author is grateful for Ludger Jansen and Stefano Borgo for their comments on the draft of the paper.

---

[4] The pattern <State → State> is not a primitive because it is explained in terms of <Process → Process>.